\documentclass{article}
\usepackage{spconf,amsmath,graphicx}
\usepackage{epsfig}
\usepackage{amssymb}
\usepackage{times}
\usepackage{wrapfig}
\usepackage{float}

\usepackage{algorithmic}
\usepackage{algorithm}

\usepackage{enumitem}
\setlist{nosep, leftmargin=14pt}

\usepackage{mwe} 

\title{Domain Adaptive Multiple Instance Learning for Instance-level Prediction of Pathological Images}
\name{
\begin{tabular}{@{}c@{}}
Shusuke Takahama$^{1}$ \qquad Yusuke Kurose$^{1,2}$ \qquad Yusuke Mukuta$^{1,2}$ \qquad Hiroyuki Abe$^{1}$ \\ 
Akihiko Yoshizawa$^{3}$ \qquad Tetsuo Ushiku$^{1}$ \qquad Masashi Fukayama$^{4}$ \qquad Masanobu Kitagawa$^{5}$ \\
Masaru Kitsuregawa$^{1,6}$ \qquad Tatsuya Harada$^{1,2,6}$
\end{tabular}
}

\address{$^{1}$ The University of Tokyo, Tokyo, Japan \\ $^{2}$ RIKEN Center for Advanced Intelligence Project, Tokyo, Japan \\ $^{3}$ Kyoto University, Kyoto, Japan  \\ $^{4}$ The Japanese Society of Pathology, Tokyo, Japan \\ $^{5}$ Tokyo Medical and Dental University, Tokyo, Japan \\ $^{6}$ National Institute of Informatics, Tokyo, Japan}
\begin{document}
%
\maketitle

\begin{abstract}
Pathological image analysis is an important process for detecting abnormalities such as cancer from cell images. However, since the image size is generally very large, the cost of providing detailed annotations is high, which makes it difficult to apply machine learning techniques. One way to improve the performance of identifying abnormalities while keeping the annotation cost low is to use only labels for each slide, or to use information from another dataset that has already been labeled. However, such weak supervisory information often does not provide sufficient performance. In this paper, we proposed a new task setting to improve the classification performance of the target dataset without increasing annotation costs. And to solve this problem, we propose a pipeline that uses multiple instance learning (MIL) and domain adaptation (DA) methods. Furthermore, in order to combine the supervisory information of both methods effectively, we propose a method to create pseudo-labels with high confidence. We conducted experiments on the pathological image dataset we created for this study and showed that the proposed method significantly improves the classification performance compared to existing methods.
\end{abstract}

\begin{figure}[!ht]
\begin{center}
\includegraphics[width=0.80\linewidth]{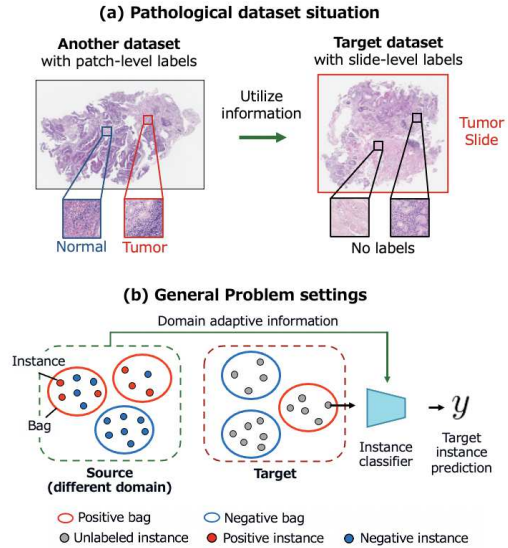}
\end{center}
\vspace{-8mm}
\caption{(a) Our problem setting. We estimate the patch labels of the target dataset only with slide-level labels. We utilize information from another dataset that already has patch-level labels. (b) The generalized setting. A WSI can be represented as a bag, which is a set of instances. Our goal is to predict target instance labels by leveraging source information.}
\label{fig:setting}
\vspace{-5mm}
\end{figure}

\section{Introduction}
\label{sec:intro}

In pathological diagnoses, doctors observe tissue slide images with a microscope and identify the presence of diseases such as cancer. Many studies have attempted to apply image recognition technology to reduce the burden on doctors through automatic diagnosis \cite{camelyon17,pathsurvey}. Because the diagnosis requires detailed cell-level observation, the size of whole slide images (WSIs) can be as large as $10^5 \times 10^5$ pixels. Owing to memory limitations, the WSI is often divided into small patch images to input classification models. Patch-level annotation takes a very high cost because it requires the expertise of doctors and a significant amount of time to annotate large WSIs. On the other hand, a label per slide, which indicates whether an abnormality exists in the WSI, requires little additional annotation cost. It is beneficial to improve the patch-level classification performance of the WSIs only with slide-level labels.

Multiple instance learning (MIL) is a type of weakly supervised learning with a single label for a bag of instances \cite{milservey,revisitmil,admil}. MIL methods have been applied to pathological image analysis, regarding the patch image as the ``instance'' and the whole slide as the ``bag'' \cite{milpath,medmil_cvpr2020}. Although this is very effective in reducing the annotation cost, the performance of the model trained only with slide labels was much lower than that with patch-level labels. On the other hand, using information from other datasets can also improve the classification performance without additional annotation costs. Domain adaptation (DA) is a method that utilizes a different domain to improve the performance of target data \cite{backprop,mcdda,pfan}. In pathological analysis, we can use existing public datasets with pixel-level labels such as the Camelyon dataset \cite{camelyon17}. However, in most cases, we cannot use them directly because of the differences in body parts, appearance, and preprocesses such as tissue staining. Some studies have attempted to overcome the differences and transfer information between different pathology datasets \cite{pathda1,pathda2}, but the performance is degraded when the difference between the domains is significant. 

In this paper, we proposed a new problem setting to improve the patch-level classification performance of the target dataset only with slide labels, while utilizing information from the labeled source dataset from another domain (Fig. \ref{fig:setting}). However, since the supervised information from the source and target dataset is qualitatively different, there is no guarantee that simply combining the two will improve performance. Therefore, we propose a new training pipeline using pseudo-labels with high reliability by combining information from both the source and target. Our method can improve performance in situations where MIL alone and DA alone cannot provide accurate classification. We performed experiments on a new pathological dataset we created for this study, and the results confirmed that our method improves the instance classification performance compared to existing methods.

\section{Method}
\label{sec:method}

\noindent \textbf{Problem settings:} In this study, we can access the labels of all instances from the source dataset, while we can only refer to bag labels and cannot access any instance labels of the target dataset. The purpose of this study is to estimate the instance labels of the target domain with high accuracy (Fig. \ref{fig:setting} (b)). In the standard MIL setting, we consider bag $X = \{{\bf x}_1, ... ,{\bf x}_K\}$ as a set of instances ${\bf x} \in \mathbb{R}^D$. $K$ is the number of instances in the bag, and it varies for each bag. Each instance has a binary label $y_k \in \{1, 0\}$, but this label cannot be referred to during training. The bag label $Y = 1$ if the bag contains at least one positive instance, and $Y = 0$ if instances are all negative. We can say that the source is a fully supervised setting, whereas the target is a standard MIL setting. We define the source domain $D_s = \{(X_i^s, Y_i^s)\}_{i=1}^{n_{sb}}$ as a set of $n_{sb}$ bags, where the $i$-th bag $X_i^s = \{({\bf x}_{ij}^s, y_{ij}^s)\}_{j=1}^{n_{si}}$ consists of $n_{si}$ labeled instances, and the target domain $D_t = \{(X_i^t, Y_i^t)\}_{i=1}^{n_{tb}}$ as a set of $n_{tb}$ bags, where the $i$-th bag $X_i^t = \{{\bf x}_{ij}^t\}_{j=1}^{n_{ti}}$ consists of $n_{ti}$ unlabeled instances. 

\noindent \textbf{Overview of our method:} Our pipeline consists of three components: encoder $G$, bag classifier $F_B$, and instance classifier $F_I$. Each instance in the bag of the target and source ${\bf x}_{ij}$ is input to $G$ to obtain the feature vectors ${\bf h}_{ij}$. The feature vectors in the bag are collectively input to the bag classifier $F_B$ to obtain the binary prediction score of the bag label $p(Y|X_i)$. By contrast, a feature vector from each instance is input to the instance classifier $F_I$ to obtain the binary prediction score of the instance label $p(y|{\bf x}_{ij})$. Because the target does not have an instance label, the source instances with the instance labels and the target instance with the pseudo labels are used for training $F_I$. At the time of inference, we input the target instance features into $F_I$ to obtain the prediction scores of the target instances $p(y|{\bf x}_{ij}^t)$.

\vspace{-5mm}
\begin{align}
    {\bf h}_{ij} &= G({\bf x}_{ij}) \\
    p(Y|X_i) &= F_B({\bf h}_{ij | j = 1...n_i}) \\
    p(y|{\bf x}_{ij}) &= F_I({\bf h}_{ij})
\end{align}



AttentionDeepMIL \cite{admil} is used as $F_B$. In this method, the bag feature is a weighted sum of instance features, and its weight $a_{ij}$ is learnable. As mentioned in \cite{admil}, the attention weight implies the positive score of each instance, so we used sigmoid instead of softmax to calculate $a_{ij}$ so that we can directly obtain the positive score of an instance: $a_{ij} = {\text {sigmoid}} ({\bf w}^T \tanh ({\bf V} {\bf h}_{ij}^T))$. ${\bf w}$ and ${\bf V}$ are hyperparameters.

To improve the prediction performance of the target instances by $F_I$, we add a domain adaptation loss that performs distribution matching of the intermediate features $h_{ij}$. We use MCD \cite{mcdda} as the DA loss. MCD performs feature distribution matching while considering category information by training the features to be away from the class boundary. To introduce MCD loss into our method, we use two instance classifiers $F_{I1}$ and $F_{I2}$. We train $G$, $F_{I1}$ and $F_{I2}$ to minimize the instance classification loss $L_I({\bf x}, y)$. At the same time, we train $G$ to minimize the discrepancy loss $L_{adv}({\bf x}^t)$ and two classifiers to maximize $L_{adv}({\bf x}^t)$ alternately. The discrepancy loss is defined as  $L_{adv}({\bf x}^t) = \frac{1}{C} \sum_{i=1}^C |p_{1i} - p_{2i}|$, where $p_1(y|{\bf x}^t)$ and $p_2(y|{\bf x}^t)$ are the output of the two classifiers.

\begin{figure}[!t]
\begin{center}
\includegraphics[width=0.90\linewidth]{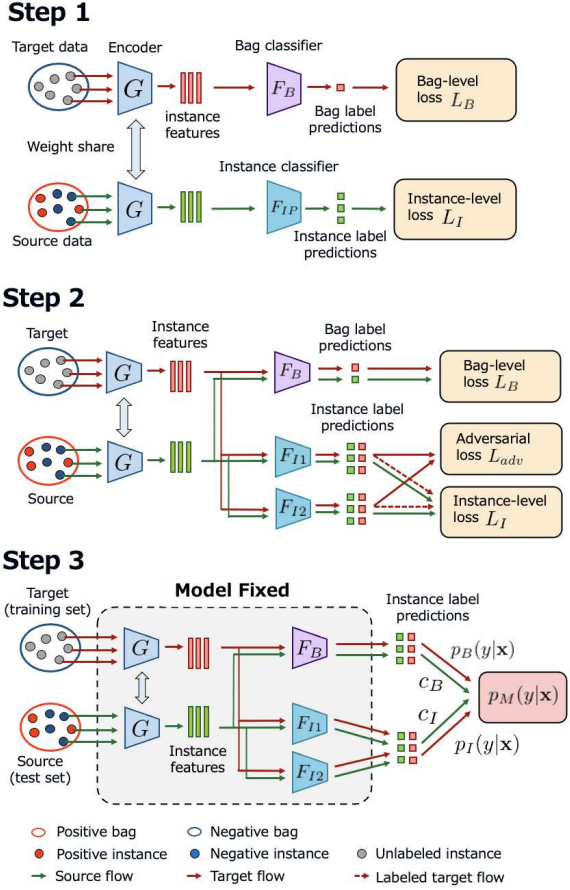}

\vspace{-3mm}
\caption{The training pipeline. Step 1 is training with $F_B$ and $F_I$ separately. After converging Step 1, we alternately perform Steps 2 and 3. Step 2 involves training with DA loss and pseudo-labeled target instances. In Step 3, we assign pseudo-labels to target instances based on $p_M(y|{\bf x})$.}
\end{center}
\label{fig:pipeline2}
\vspace{-5mm}
\end{figure}

\noindent \textbf{Pseudo labeling:} Even if the feature distributions match, there will still be many misclassified instances if the decision boundaries of the source and target do not match. To tackle this problem, we directly optimize our model for target instance prediction by assigning pseudo-labels to the target instances and using them for the training of $F_I$. Because we know that all instances in the negative bag are negative, we mainly consider the instances from the positive bags. To obtain reliable pseudo-labels, we use two classifiers $F_B$ and $F_I$. Because these two classifiers are trained using different supervisory information, they have different properties. we can obtain pseudo-labels with higher reliability by integrating the information from both of them.

We define the prediction score of the instance classifier $p_I(y|{\bf x})$ as the average of the predictions of $F_{I1}$ and $F_{I2}$. We can also obtain the instance prediction score of the bag classifier $p_B(y|{\bf x})$ using the attention weight $a_{ij}$ of $F_B$. Because the two classifiers are trained in different ways, the accuracy of the predictions of both models can vary. For example, if the prediction performance of $F_B$ is significantly poor, then the prediction of $F_I$ should be mainly used. Therefore, we consider the confidence score of each model when assigning pseudo-labels. $c_B$ and $c_I$ represent the confidence scores of $F_B$ and $F_I$, respectively. Because the target instances have no labels and we cannot directly examine the prediction accuracy, we instead use the PR-AUC score of the prediction performance of the source instances by each model as the confidence score. Then, we define the mix prediction score $p_M(y|{\bf x})$ as:
\begin{align}
\label{eq:mixpred}
    p_M(y|{\bf x}^t) = c_B * p_B(y|{\bf x}^t) + c_I * p_I(y|{\bf x}^t)
\end{align}

We first select positive candidates that satisfy $p_M(y = 0|{\bf x}^t) \leq p_M(y = 1|{\bf x}^t)$ and $0.5 \leq p_M(y = 1|{\bf x^t})$, and negative candidates that satisfy $p_M(y = 1|{\bf x}^t) \leq p_M(y = 0|{\bf x}^t)$ and $0.5 \leq p_M(y = 0|{\bf x^t})$. Then, from each of the positive and negative candidates, we select a fixed number of instances that have high confidence scores and assign pseudo-labels to them. We give pseudo-labels only to a small number of reliable instances at the beginning of the training when the prediction is ambiguous, and gradually increase the number as the training progresses. The definition of the number of pseudo-labels appears in the supplementary materials.

\noindent \textbf{Training process:} Figure \ref{fig:pipeline2} shows the entire pipeline of our method. Our method consists of three steps. In Step 1, we perform supervised learning of $F_B$ using the target bag labels and $F_I$ using the source instance labels. In this case, we use only one instance classifier $F_{IP}$. By performing Step 1, the training becomes more stable, and reliable pseudo-labels can be obtained from the beginning of Step 3. After Step 1 converges, we initialize two instance classifiers $F_{I1}$ and $F_{I2}$ and perform Steps 2 and 3 alternately. In Step 2, we train the model using the feature matching loss of DA. $F_B$ is trained using both the source and target data. In addition, $F_{I1}$ and $F_{I2}$ are trained with $\{{\bf x}^m, y^m\}$, which includes the source instances, the target instances with pseudo-labels from positive bags, and the sampled target instances from negative bags. We optimize the following three losses individually:

\footnotesize
\vspace{-3mm}
\begin{align}
    \underset{G, F_{I1}, F_{I2}, F_B}{\min} & \lambda L_I({\bf x}^m, y^m) + (1-\lambda) (L_B(X^t, Y^t) + L_B(X^s, Y^s)) \label{eq:loss1}\\
    \underset{F_{I1}, F_{I2}}{\min} ~~&\lambda(L_I({\bf x}^m, y^m) - L_{adv}({\bf x}^t)) \label{eq:loss2}\\
    \underset{G}{\min} ~~&\lambda L_{adv}({\bf x}^t) \label{eq:loss3}
\end{align}
\normalsize

\noindent where $\lambda$ is a weight parameter. In Step 3, we fix the model parameters and give pseudo-labels to the target instances from positive bags. At the same time, we input the source instances in the validation set to calculate $c_B$ and $c_I$ in (\ref{eq:mixpred}).

  %
  %
\begin{table}
\begin{center}
\caption{The classification performance of each method on pathological dataset}
  \small
  \begin{tabular}{c|c c} \hline
     & Accuracy & PR-AUC\\ 
    \hline \hline
    Attention MIL & 72.4$\pm$5.62 & 66.0$\pm$1.24 \\
    Source only & 82.5$\pm$1.20 & 66.9$\pm$1.00 \\
    MCDDA & 76.0$\pm$3.09 & 51.8$\pm$2.78 \\
    PLDA & 78.6±1.66 & 76.5$\pm$3.97 \\ \hline
    Ours (Step 1) & 82.7$\pm$2.84 & 71.1$\pm$2.62 \\
    Ours & {\bf 86.0±4.11} & {\bf 83.4$\pm$3.48} \\ \hline
    Ideal case & 91.1$\pm$0.79 & 87.1$\pm$2.05 \\ \hline
  \end{tabular}
\label{tb:result_path}
\vspace{-8mm}
\end{center}
\end{table}
\section{Experiment}
\label{sec:Experiment}

In this section, we present the experimental results to confirm the effectiveness of our method. As a preliminary experiment, we performed detailed evaluations and ablation studies using benchmark datasets. The details are provided in the supplementary materials. In the following, we describe the results of the experiments using our pathological dataset.

\noindent \textbf{Dataset:} We constructed a new original dataset of pathological images to demonstrate the effectiveness of the proposed method. We collected whole slide images (WSI) from two body parts, ``Stomach'' and ``Colon,'' which include 997 and 1368 WSIs, respectively.
Many previous studies have set WSIs of two different datasets from a single organ as the source and the target, respectively \cite{pathda1,pathda2}. However, the domain gap between two organs is considerably larger than that in a single organ, making our settings more challenging and suitable for demonstrating the effectiveness of our method.

The size of the WSIs is approximately $10^4 \times 10^4$ pixels, and the maximum resolution is $\times20$. Figure \ref{fig:setting} (a) shows an example of the stomach WSI (right) and the colon WSI (left). The pixel-level normal/abnormal annotation was provided by expert pathologists. We separated WSI into patches of size $256 \times 256$ without overlaps and assign a binary label to each patch based on pixel-level annotation. The cropping, labeling, and image pre-processing methods followed the approach in \cite{takahama}. We use Colon as the source and Stomach as the target.

Next, we created bags from each slide. Because one slide contains, at most, several hundred patches, we sampled 30 patches to make one bag. For the target positive slide, there's no guarantee a bag includes positive instance because we don't have patch labels. To increase the probability that the bag contains a positive instance, we performed clustering. First, we obtained the features of target patches using a classifier trained with source instances. Then, we separated the features into 10 clusters using K-means, and selected three samples with high positive scores from each cluster to obtain a bag of size 30. As a result, the probability that at least one positive instance is included in a bag created from the positive slide (confidence level of positive bag labels) was 95.0\%, which is sufficiently reliable. For the negative slide, because all the patches were negative, we made as many bags as possible by randomly selecting patches. We randomly separated our dataset into 70\% training slides and 30\% test slides. Finally, we obtained 1000 training bags and 200 test bags from the source and target dataset, respectively.



\begin{figure}[t]
\begin{center}
\includegraphics[width=1.0\linewidth]{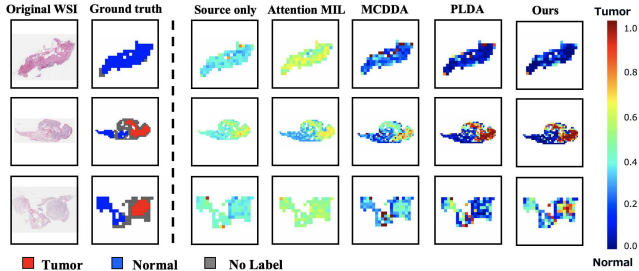}
\end{center}
\vspace{-4mm}
\caption{Prediction heatmaps of positive prediction scores for target stomach dataset. In the ground-truth map, red indicates positive (anomaly), blue indicates negative (normal), and gray indicates areas without annotation.}
\label{fig:pathology_heatmap}
\vspace{-5mm}
\end{figure}

\noindent \textbf{Comparison methods:} Since our problem setting is completely new, it cannot be compared to SOTA methods directly. To verify the effectiveness of MIL, DA, and pseudo-label modules, we evaluate the following comparison methods.

\begin{itemize}
  \setlength{\parskip}{0cm}
  \setlength{\itemsep}{0cm}
  \item \textbf{Attention MIL}: Train AttentionDeepMIL \cite{admil} only with the target data. We used the values of the attention weights as the instance prediction scores.
  \item \textbf{Source only}: Train $G$ and $F_I$ with only source instances.
  \item \textbf{MCDDA}: Train unsupervised DA pipeline of MCD \cite{mcdda} without the bag labels of target data.
  \item \textbf{PLDA}: Train unsupervised DA with pseudo-labels. We assign pseudo-labels to the target instances by the model trained with the source and use them for the training.
  \item \textbf{Ours (Step 1)}: Train only Step 1 of the proposed method. We evaluate the classification performance of $F_{IP}$.
  \item \textbf{Ideal case}: Train $G$ and $F_I$ using target instance labels that are not actually available. This is considered as the upper bound of the classification performance.
\end{itemize}
\vspace{3mm}

\noindent \textbf{Experimental Settings:} We used accuracy and the AUC of the precision-recall curve (PR-AUC) for instance-level prediction as the evaluation metrics. All experiments were conducted three times with random initial model weights, and the mean and standard deviation were calculated. We used ResNet50 \cite{resnet} pretrained with ImageNet \cite{imagenet} as $G$, and two fully connected layers as $F_I$. The dimension of the output of $G$ was 500. We trained 50 epochs for pretraining and source only, and 100 epochs for others. We set $\lambda = 0.5$.

\noindent \textbf{Results:} Table \ref{tb:result_path} shows the results of each method. Our proposed method outperforms other methods and achieves comparable scores with ``Ideal case.'' Figure \ref{fig:pathology_heatmap} shows heatmaps of the estimated patch labels in the WSIs in the target test set by each trained model. The heatmaps of ``Source only'' and ``Attention MIL'' show little difference between the scores of the normal and abnormal areas, which implies that the predictions appear relatively vague. The heatmaps of ``MCDDA'' and ``PLDA'' appear to be relatively reasonable, but there are some regions of high abnormality scores in the normal region. This result is unfavorable for practical purposes because doctors need to examine the slide even if there is a small abnormality area. And ``MCDDA'' and ``PLDA'' do not detect the abnormal region well in the bottom example. Our proposed method made qualitatively valid prediction maps with a clear difference between the prediction scores of the normal and abnormal regions. Our method proved to be effective even in real-world applications such as pathological images.

\section{Conclusion}
\label{sec:Conclusion}

In this study, we proposed a new problem setting to improve the classification performance of pathological images with low annotation cost, using only slide-level labels and information of another dataset from a different domain. In addition, we proposed a new pipeline to achieve the accurate classification of target instances by assigning pseudo-labels using two different supervisory information. Our method was evaluated on the pathological image dataset constructed in this study. The results demonstrate that our proposed method can achieve higher performance than comparative methods.

\vfill
\pagebreak

\section{Acknowledgments}
This work was partially supported by AMED JP18lk1010028 $\cdot$ JP19lk1010036, JST AIP Acceleration Research JPMJCR20U3, Moonshot R\&D Grant Number JPMJPS2011, CREST Grant Number JPMJCR2015, JSPS KAKENHI Grant Number JP19H01115 $\cdot$ JP19K20369 and Basic Research Grant (Super AI) of Institute for AI and Beyond of the University of Tokyo.
\section{Compliance with Ethical Standards}
This study was performed in line with the principles of the Declaration of Helsinki. Approval was granted by the Ethics Committee of The University of Tokyo (9/1/2021, 21-222)
\bibliographystyle{IEEEbib}
\bibliography{main}

\end{document}


%
\maketitle

\section{Method}
\subsection{Number of pseudo-labels}
We give pseudo-labels only to a small number of reliable instances at the beginning of the training when the prediction is ambiguous, and gradually increase the number as the training progresses. $\tau$ is a variable for the upper limit of the number of pseudo-labels:
\begin{align}
    \tau = \min \left(N_{min} + m * \frac{N_{max} - N_{min}}{M} , N_{max} \right)
\end{align}

\noindent where $m$ denotes the training epoch, and $\tau$ increases depending on $m$. We assign labels to $A_p * \tau$ positive instances and $A_n * \tau$ negative instances. $M$, $N_{min}$, $N_{max}$, $A_p$ and $A_n$ are the hyperparameters. For the target negative bags, all instances are guaranteed to be negative. To maintain a balance between the number of labeled instances from the positive and negative bags, instances were randomly sampled from negative bags to equal the number of labeled instances from the positive bags.

We set $M = 20$, $A_p = A_n = 1$, $N_{min} = \frac{n_{tpi}}{10}$, and $N_{max} = \frac{n_{tpi}}{4}$, where $n_{tpi}$ is the sum of the instances in all target positive bags in the experiment with the pathological dataset. 

\section{Experiments on digit datasets}

\subsection{Details} 
In this supplementary material, we present the results of preliminary experiments to confirm the effectiveness of our method. We performed detailed evaluations using digit datasets (MNIST \cite{mnist} and SNHN \cite{svhn}) and VisDA dataset \cite{visda}, a benchmark datasets for image analysis and domain adaptation.

First, we describe the experiments using the digit datasets. We created a bag by collecting images from MNIST \cite{mnist} and SVHN \cite{svhn}, as in \cite{admil}. We define ``9'' as a positive class and the others as a negative class. The bag label is positive when at least one ``9'' is in it, and otherwise negative. The bag size follows a normal distribution with a mean of 10 and a variance of 2, and the number of positive instances in a positive bag follows a normal distribution with a mean of 1 and a variance of 1. In both datasets, we made 2000 training bags and 500 test bags, including an equal number of positive and negative bags by random sampling.

We used an encoder with three convolution layers, followed by a fully connected layer as $G$. $F_I$ consists of two fully connected layers. The size of the output of the encoder was 500. We use Adam \cite{adam} as the optimizer, and the learning rate was 1e-4. We trained 50 epochs for pre-training and source only, and 100 epochs for others. We set $\lambda = 0.5$, $A_p = 1$, $A_n = 3$, $N_{min} = \frac{n_{tpi}}{30}$, and $N_{max} = \frac{n_{tpi}}{10}$, where $n_{tpi}$ is the sum of the instances in all target positive bags. Other settings and model structures were the same as the experiment on the pathological dataset.

\begin{table*}
\begin{center}
\caption{The classification performance on VisDA dataset.}
\scalebox{0.94}{
  \begin{tabular}{c|c c c c c c c c c c c c|c} \hline
     & plane & bcycl & bus & car & horse & knife & mcycl & person & plant & sktbrd & train & truck & mean\\ 
    \hline \hline
    Attention MIL & 95.1 & 79.6 & 81.4 & 5.5 & 5.0 & 89.3 & 6.4 & 5.6 & 89.5 & 68.2 & 6.3 & 6.8 & 44.9 \\
    Source only & 77.4 & 50.9 & 46.9 & 24.0 & 60.5 & 25.2 & 59.6 & 35.6 & 59.1 & 29.9 & 39.3 & 22.6 & 44.3 \\
    MCDDA & 91.4 & 75.1 & 75.5 & 43.5 & 89.7 & 24.2 & 80.4 & 43.8 & 80.3 & 44.2 & 76.7 & 41.7 & 63.9 \\
    PLDA & 96.9 & 86.0 & 79.5 & 52.2 & 94.1 & 89.0 & 86.2 & 64.2 & 89.9 & 70.4 & 84.4 & 56.4 & 79.1 \\ \hline
    Ours (Step 1) & 92.2 & 75.0 & 78.3 & 40.1 & 87.5 & 28.3 & 83.4 & 47.5 & 85.0 & 50.1 & 75.5 & 40.5 & 65.3 \\
    Ours & {\bf 98.3} & {\bf 88.1} & {\bf 84.7} & {\bf 59.5} & {\bf 95.5} & {\bf 93.7} & {\bf 88.4} & {\bf 73.6} & {\bf 94.2} & {\bf 80.1} & {\bf 89.8} & {\bf 60.6} & {\bf 83.9} \\ \hline
    Ideal case & 99.0 & 91.4 & 88.7 & 67.5 & 96.2 & 96.6 & 90.3 & 83.4 & 95.6 & 93.0 & 93.2 & 71.3 & 88.9 \\ \hline
  \end{tabular}
  }
\label{tb:result_visda}
\vspace{-5mm}
\end{center}
\end{table*}

\begin{table}
\begin{center}
\caption{The classification performance on the digit dataset.}
  \begin{tabular}{c|c c} \hline
    Source & SVHN & MNIST\\ 
    Target & MNIST & SVHN\\ 
    \hline \hline
    Attention MIL & {\bf 99.5$\pm$0.39} & 24.1$\pm$1.13 \\
    Source only & 37.4$\pm$1.09 & 14.6$\pm$1.46 \\
    MCDDA & 11.5$\pm$1.54 & 9.5$\pm$1.20 \\
    PLDA & 98.0$\pm$1.81 & 53.6$\pm$1.51 \\ \hline
    Ours (Step 1) & 97.1$\pm$0.56 & 30.6$\pm$2.27 \\
    Ours & {\bf 99.4$\pm$0.56} & {\bf 80.7$\pm$0.56} \\ \hline
    Ideal case & 99.8$\pm$0.00 & 86.7$\pm$0.29 \\ \hline
  \end{tabular}
\label{tb:result_digit}
\vspace{-5mm}
\end{center}
\end{table}

\subsection{Result}
Table \ref{tb:result_digit} lists the prediction scores for each method. In the setting where MNIST is set as the target, our method shows high performance, but the improvement is negligible because ``Attention MIL'' and ``PLDA'' already yielded good predictions. The reasons for the poor ``MCDDA'' performance may be an imbalance in the number of positive and negative instances and the lack of pretraining with ImageNet \cite{imagenet} of the model. On the other hand, in the setting where SVHN is set as the target, the proposed method performs considerably better than the comparative methods, because SVHN is more difficult to classify than MNIST. Our method achieves PR-AUC close to the ``Ideal case," owing to the complementary effect of the target bag-level loss and source instance-level loss. We found that our method achieved a high classification performance without additional annotation of the target instances.

\section{Experiments on VisDA 2017}
\subsection{Details} In this section, we validate our proposed method using VisDA 2017 \cite{visda}. VisDA 2017 is one of the largest cross-domain datasets for object classification and consists of a training dataset of synthetic images, a validation dataset of real images collected from MSCOCO \cite{mscoco}, and a test dataset of real images collected from a different domain from the validation dataset. Because the label of the test dataset is not available, we used the training dataset as the source domain and the validation dataset as the target domain. Because there are 12 categories, we evaluated the performance when one category was set as positive and the others as negative for every category.

In each setting, we created 500 training bags and 200 test bags. The bag size follows a normal distribution with a mean of 10 and a variance of 2. The number of positive instances in a positive bag follows a normal distribution with a mean of 2 and a variance of 2. The learning rate is 1e-6 for methods including feature distribution matching loss and 1e-5 for others. We trained 50 epochs for pretraining and source only, and 100 epochs for others. Other settings and model structures were the same as the experiment on the pathological dataset.

\begin{table}
\begin{center}
\caption{The result of the ablation study on VisDA dataset.}
\scalebox{0.92}{
  \begin{tabular}{c|c c c c c} \hline
     & bus & car & knife & sktbrd\\ 
    \hline \hline
    Ours (Step 1) & 78.3 & 40.1 & 28.3 & 50.1 \\ \hline
    w/o pseudo-label & 79.2 & 52.3 & 51.2 & 57.9 \\
    w/o feature matching & 84.0 & 57.5 & 92.7 & 76.4 \\
    pseudo-label with $F_I$ & 84.2 & 59.1 & 93.3 & 78.9 \\
    pseudo-label with $F_B$ & {\bf 84.9} & 15.3 & {\bf 93.8} & 11.5 \\
    pseudo-label w/o conf. score & 84.6 & 57.5 & 93.5 & 78.3 \\
    pseudo-label with PFAN & 84.0 & 57.1 & 91.7 & 76.7 \\ \hline
    Ours & {\bf 84.7} & {\bf 59.5} & {\bf 93.7} & {\bf 80.1} \\ \hline
  \end{tabular}
  }
\label{tb:result_ablation}
\vspace{-1mm}
\end{center}
\end{table}

\subsection{Results}
Table \ref{tb:result_visda} shows the performance of our proposed method and the comparison methods when each category is set as positive. Our method performed better than the comparison methods in all settings. In particular, our method improves the performance of categories such as ``car'' and ``truck,'' which are easily mistaken for other vehicle categories and show extremely low performance with ``Attention MIL.'' Further, our method shows comparable scores with the ``Ideal case'' when ``plane'' and ``horse'' are set as positive. In addition, ``Ours (Step 1),'' which simply combines DA and MIL, does not achieve high performance, which demonstrates the effectiveness of our method.

\begin{figure*}[t]
\begin{center}
\includegraphics[width=1.0\linewidth]{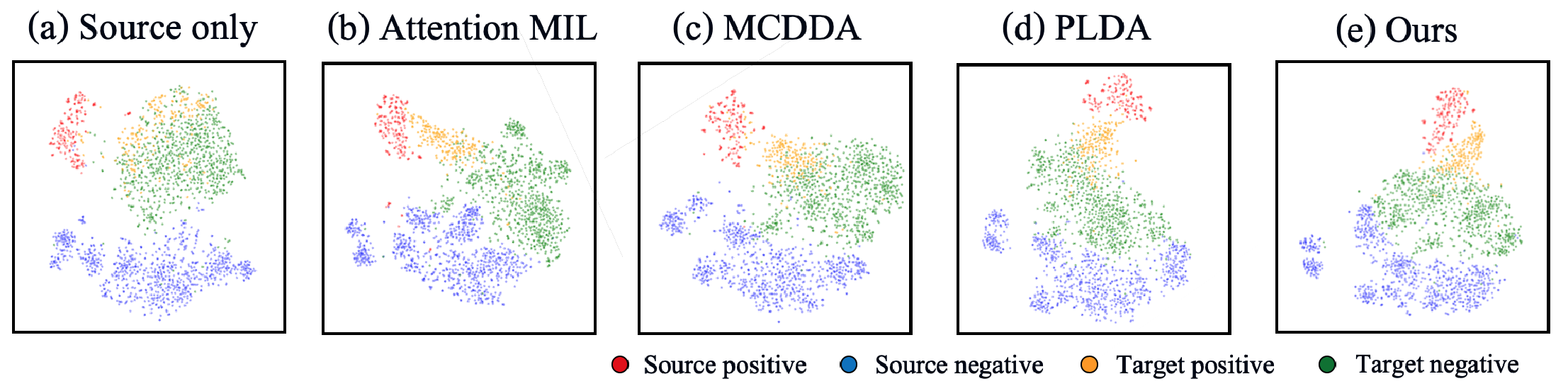}
\end{center}
\vspace{-5mm}
\caption{Visualization of the feature of VisDA dataset when ``plant'' is positive. Red and blue indicate the source positive and negative instances, respectively, and yellow and green indicate the target positive and negative instances, respectively. We can observe that our proposed method achieves feature distribution matching and obtains a discriminative decision boundary.}
\label{fig:feature}
\vspace{3mm}
\end{figure*}

\begin{figure*}[t]
\begin{center}
\includegraphics[width=1.0\linewidth]{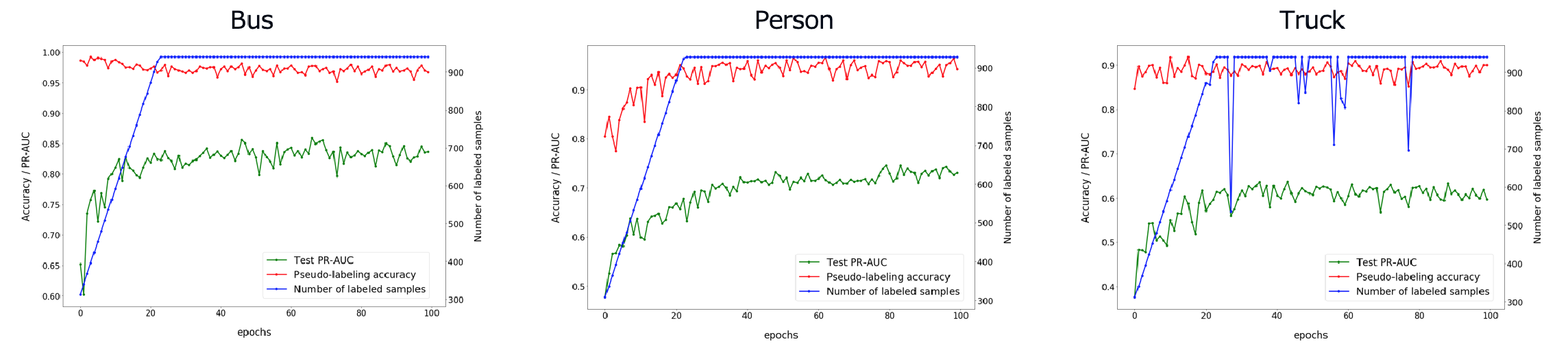}
\end{center}
\vspace{-5mm}
\caption{Comparison of the number of labeled target instances, actual accuracy of pseudo-labels, and PR-AUC score during training on VisDA dataset.}
\label{fig:label_plot}
\end{figure*}

\subsection{Feature distribution} 
Figure \ref{fig:feature} presents the visualization result for the distribution of intermediate features after training each method when the ``plant'' is set as positive. In ``Source only,'' the decision boundary of the target negative and positive are ambiguous, and the distribution of the source and target are completely separated. Although the separation of the target positive and negative is slightly improved in ``MCDDA'' and ``PLDA,'' it is not sufficient for accurate classification. Moreover, in ``Attention MIL,'' although the decision boundary of the target is clearer, the target distribution is completely separated from the source, and there is no guarantee that the target data can be successfully classified by the decision boundary of $F_I$. In our proposed method, the decision boundary of the target is clear, and the distributions of the source and target are well-matched. This means that $F_I$ can achieve good classification performance, even in target instances.

\subsection{Accuracy of pseudo-labels}
Figure \ref{fig:label_plot} shows the number of target instances given pseudo-labels, the actual accuracy of pseudo-labels, and the PR-AUC score of the proposed method for each training epoch on the VisDA dataset. The figure shows the results when ``bus,'' ``person,'' and ``truck'' are set as positive. The upper limit of the number of labeling instances was gradually increased up to the 20th epoch, and in most cases, the upper limit number of instances was labeled. Sometimes, as in the case of the ``truck,'' the number of labeled instances was lower than the upper limit, but the performance of the model improved steadily even in such cases. In the case of ``bus,'' although the pseudo-label accuracy decreased slightly as the number of labeled instances increased, it was maintained at a high level. By contrast, in the case of ``person,'' the pseudo-label accuracy increased as the model performance improved.

\subsection{Ablation study}
\label{sec:ablation}
In our proposed method, the predictions of $F_B$ and $F_I$ are combined based on the confidence scores of the model to obtain $p_M(y|{\bf x^t})$, and pseudo-labels are assigned depending on the $p_M(y|{\bf x^t})$ score. In addition, we added DA loss to achieve feature distribution matching. To clarify the effect of each module, we conducted ablation studies using the following methods:

\begin{itemize}
  \setlength{\parskip}{0cm}
  \setlength{\itemsep}{0cm}
  \item \textbf{w/o pseudo-label}: Train without pseudo-labels.
  \item \textbf{w/o feature matching}: Train without feature matching loss.
  \item \textbf{pseudo-label with $F_I$}: Assign pseudo-labels considering only the prediction of $F_I$.
  \item \textbf{pseudo-label with $F_B$}: Assign pseudo-labels considering only the prediction of $F_B$.
  \item \textbf{pseudo-label w/o conf. score}: Assign pseudo-labels without considering confidence scores $c_I$ and $c_B$. We use the simple sum $p_M(y|{\bf x}^t) = p_B(y|{\bf x}^t) + p_I(y|{\bf x}^t)$.
  \item \textbf{pseudo-label with PFAN}: Giving pseudo-labels in feature space as in PFAN \cite{pfan}, which is one of the SOTA methods of pseudo-label DA. Specifically, we calculate the positive and negative centroids of the source instances and assign pseudo-labels to the target instances that are close to the centroids.
\end{itemize}

Table \ref{tb:result_ablation} shows the results of evaluating each method when ``bus,'' ``car,'' ``knife,'' and ``skate'' are set as positive. First, the performance of ``w/o pseudo-label'' and ``w/o feature matching'' is degraded, indicating that pseudo-labels and feature matching loss contribute to the performance improvement. With respect to ``pseudo-label with $F_B$,'' the performance for ``bus'' and ``knife'' is good, while the performance for ``car'' and ``truck'' is remarkably poor. This result is owing to the low confidence of $F_B$, which could not provide accurate pseudo-labels. By contrast, if the performance of $F_I$ is extremely poor, then the performance of ``the pseudo-label with $F_I$'' becomes worse. This indicates that the proposed method, which integrates the scores of both $F_I$ and $F_B$ considering the confidence score, contributes to performance improvement. In addition, the performance of ``pseudo-label w/o conf. score,'' which uses a simple sum of two predictions, is inferior to the proposed method. Furthermore, the proposed method outperforms PFAN which uses only a single score from feature space for labeling. This confirms the effectiveness of considering information from two models with different properties.

\vfill
\pagebreak

\bibliographystyle{IEEEbib}
\bibliography{supple}